\documentclass[runningheads,a4paper]{llncs}
\usepackage{times}
\usepackage{url}
\usepackage{latexsym}
\usepackage{graphicx}
\usepackage{multicol}
\usepackage{booktabs}
\usepackage{array}
\usepackage{amssymb}

\usepackage{graphicx}
\usepackage[utf8]{inputenc}

\usepackage{multirow}
\usepackage{color}

\graphicspath{{figures/}}

\usepackage{cancel}
\usepackage[normalem]{ulem}

\newcommand\ie[1]{\textit{i.e.~}}
\newcommand\eg[1]{\textit{e.g.~}}

\title{Neural Machine Translation By Generating Multiple Linguistic Factors}

  \author{Mercedes Garc\'ia-Mart\'inez \and 
  Lo\"ic Barrault \and Fethi Bougares}

\institute{LIUM, Le Mans University, France\\
\email{FirstName.LastName@univ-lemans.fr}\\ 
}

\date{}

\begin{document}
\maketitle
\begin{abstract}
Factored neural machine translation (FNMT) is founded on the idea of using the morphological and grammatical decomposition of the words (factors) at the output side of the neural network.
This architecture addresses two well-known problems occurring in MT, namely the size of target language vocabulary and the number of unknown tokens produced in the translation.
FNMT system is designed to manage larger vocabulary and reduce the training time (for systems with equivalent target language vocabulary size). 
Moreover, we can produce grammatically correct words that are not part of the vocabulary.
FNMT model is evaluated on IWSLT'15 English to French task  and compared to the baseline word-based and BPE-based NMT systems.
Promising qualitative and quantitative results (in terms of BLEU and METEOR) are reported.

\end{abstract}

\section*{Keywords}
Machine Translation, Neural Networks, Deep Learning, Factored representation

\section{Introduction and related works}
\label{intro}

In contrast to the traditional phrased-based statistical machine translation~\cite{Koehn} that automatically translates subparts of the sentences, standard Neural Machine Translation (NMT) systems use the sequence to sequence approach at word level and consider the entire input sentence as a unit for translation~\cite{Cho,Bahdanau,SutskeverVL14}. 

Recently, NMT showed better accuracy than existing phrase-based systems for several language pairs. 
Despite these positive results, NMT systems still face several challenges. 
These challenges include the high computational complexity of the softmax function which is linear to the target language vocabulary size (Equation~\ref{eq:softmax}).
	\begin{equation} \label{eq:softmax}
	p_i =  e^{o_i} / \sum_{r=1}^N e^{o_r} ~\mathrm{for}~ i \in \{1, \dots, N\}
	\end{equation}
where $o_i$ are the outputs, $p_i$ their softmax normalization and $N$ the total number of outputs.\\

In order to solve this issue, a standard technique is to define a \emph{short-list} limited to the $s$ most frequent words where $s << N$. 
The major drawback of this technique is the growing rate of unknown tokens generated at the output. 
Another work around has been proposed in \cite{Jean} by carefully organising the batches so that only a subset $K$ of the target vocabulary is possibly generated at training time. 
This allows the system to train a model with much larger target vocabulary without substantially increasing the computational complexity. 
Another possibility is to define a structured output layer (SOUL) to handle the words not appearing in the shortlist. 
This allows the system to always apply the softmax normalization on a layer with reduced size \cite{Le}. 
The problem of unknown words was addressed making use of the alignments produced by an unsupervised aligner \cite{LuongSLVZ14}.
The unknown generated words are substituted in a post-process step by the translation of their corresponding aligned source word or copying the source word if no translation is found.
The translation of the source word is made by means of a dictionary.

Other recent work have used subword units instead of words.
In \cite{Sennrich16BPE}, some unknown and rare words are encoded as subword units with the Byte Pair Encoding (BPE) method. 
Authors show that this can also generates words unseen at training time.
As an extreme case, the character-level neural machine translation has been presented in several works~\cite{Chung,Ling15,Costa-Jussa} and showed very promising results.
The character-level NMT architectures are composed of many layers, to deal with the long distance dependencies, increasing aggressively the computational complexity of the training process. 
In \cite{Sennrich2016HowGI} has been shown that character-level decoders outperform subwords units using BPE method when processing unknown words, but they perform worse when extracting morphosyntactic information about the sentences, due to the long distances.

Among other previous works, our work can be seen as a continuation of ~\cite{Garcia16iwslt}. 
Several works have used factors as additional information for the input words in neural language modelling with interesting 
results~\cite{niehues,Alexandrescu,Wu2012FactoredRN}. 
More recently, factors have also been integrated into a word-level NMT system as additional linguistic input features~\cite{SennrichH16}.
Unlike these previous works, we are considering factors as translation unit.
We refer to \emph{factors} as some linguistic annotations at word level, \eg\ the Part of Speech (POS) tag, number, gender, etc. 
The advantages of using factors as translation unit are two-fold: 
reducing the output vocabulary size and allowing to generate surface forms which are never seen in the training data.\\

 
Factors were first introduced for NMT at output side in~\cite{Garcia16iwslt} where two factored synchronous symbols are simultaneously generated. 
Authors presented an investigation of the architecture of their factored NMT system to show that better results are obtained using a feedback of the two generated outputs concatenation.

Our work is different from previous efforts in that we consider only the best type of feedback for the network. 
We also introduce an additional factor about the case information (lowercase, uppercase or in capitals)
and evaluate using a different translation test.
Moreover, we apply an unknown words (\emph{unk}) replacement technique using the alignments of the attention mechanism to replace the generated unknown words in target side.
For that, we make use of an unigram dictionary to find the translation of the source word corresponding to the generated \emph{unk}.

We compare this architecture to the state of the art BPE approach and the classic word-level NMT approach on the English to French dataset from IWSLT'15 evaluation campaign. 
We provide, in addition a quantitative and qualitative study about the obtained results.\\

The remainder of this paper is organized as follows: Section~\ref{se:nmt} describes the  attention-based NMT system and Section~\ref{se:fnmt} its extension using the factored approach. 
In Section~\ref{se:experiments}, we describe the experiments and the obtained results. 
Finally, Section~\ref{se:conclusion} concludes the paper and presents the future work.

\section{Neural Machine Translation}
\label{se:nmt}

The standard NMT model consists of a sequence to sequence encoder-decoder of two recurrent neural networks (RNN), one used by the encoder and the other by the decoder. 
The source language sequence is mapped into an embedded dimension in the encoder and the decoder maps the representation back to a target language sequence.

\begin{figure}[!htbp]
\centerline{\includegraphics[scale=0.5]{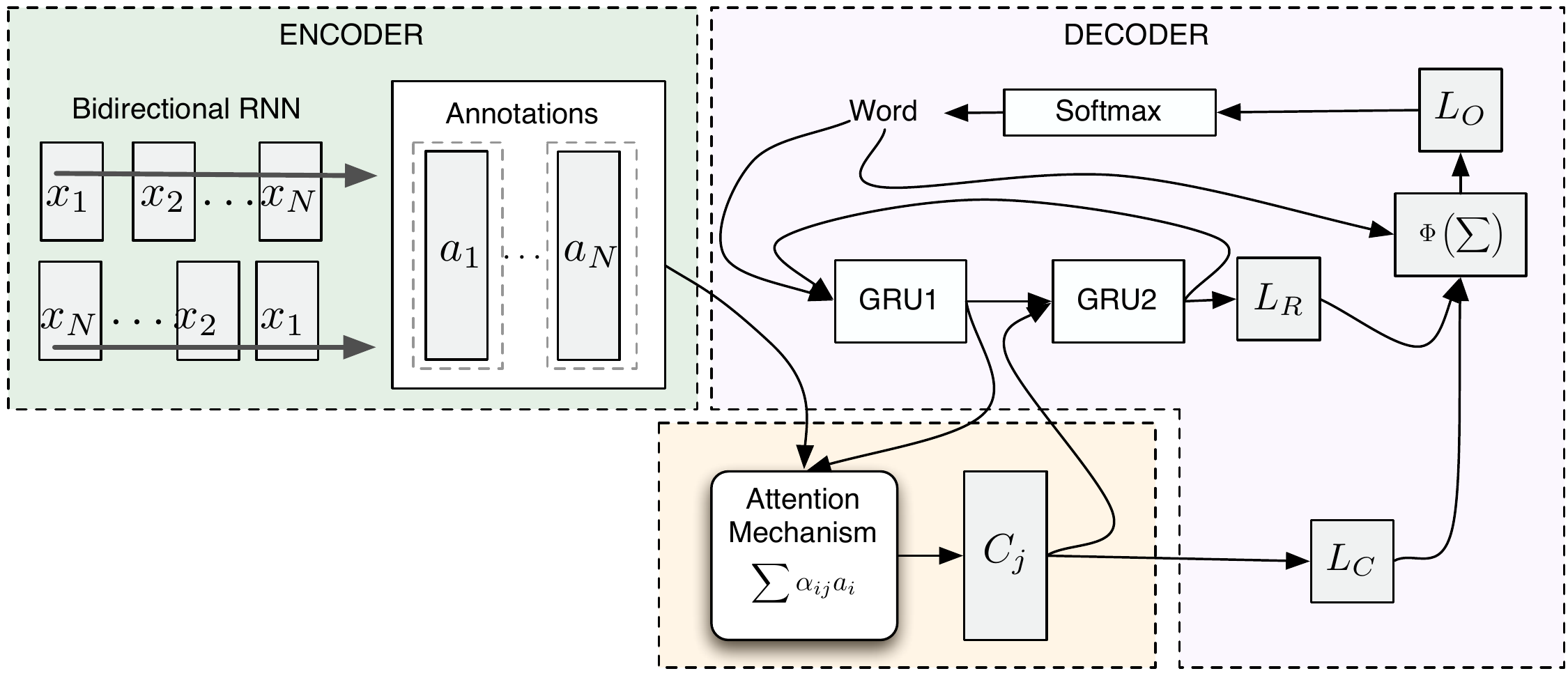}}
\caption{\label{fi:nmt_archi} Attention-based NMT system}
\end{figure}
The architecture includes a bidirectional RNN encoder (see left part of Figure~\ref{fi:nmt_archi}) equipped with an attention mechanism \cite{Bahdanau}.
Each input sentence word $x_i$ ($i \in 1\dots{}N$ with $N$ the source sequence length) is encoded into an annotation $a_i$ by concatenating the hidden states of a forward and a backward RNN provided by a gated recurrent unit (GRU) \cite{Cho} to control the flow of information. 
These annotations $a_1\dots{}a_N$ represents the whole sentence with a focus on the word being processed. 
One difference from the architecture of \cite{Bahdanau} is that the decoder contains a conditional GRU \cite{cgru} which consists of two GRUs interspersed with the attention mechanism (see right top part of the Figure~\ref{fi:nmt_archi}). 
The first GRU combines the embedding of the previous decoded token and the previous hidden state in order to generate an intermediate representation which is an input of the attention mechanism and the second GRU.
The attention mechanism (bottom yellow part of the Figure~\ref{fi:nmt_archi}) computes a source context vector $C_j$ as a convex combination of annotation vectors, where the weights of each annotation are computed locally using a feed-forward network.
These weights can be used to align the target words with the source positions.
The second GRU generates the hidden state of the conditional GRU by looking at the output of the first GRU and the context vector $C_j$.
The decoder RNN takes as input the embedding of the previous output word (feedback of the network) in the first GRU, the context vector  $C_j$ in the second GRU and its hidden state. 
The output layer $L_O$ is connected to the network through a hyperbolic tangent sum operation $\Phi (\sum$) which takes as input the embedding of the previous output word as well as the context vector and the output of the decoder from the second GRU (both adapted with a linear transformation, respectively, $L_C$ and $L_R$). 
Finally, the output probabilities for each word in the target vocabulary are computed with a \emph{softmax} function. 
The word with the highest probability is the translation output at each timestep.
The encoder and the decoder are trained jointly to maximize the conditional probability of the reference translation.

\section{Factored Neural Machine Translation}
\label{se:fnmt}

The Factored Neural Machine Translation (FNMT)~\cite{Garcia16iwslt} is an extension of the standard NMT architecture which allows the system to generate several output symbols at the same time.

\begin{figure}[!htbp]
	\centerline{\includegraphics[scale=0.6]{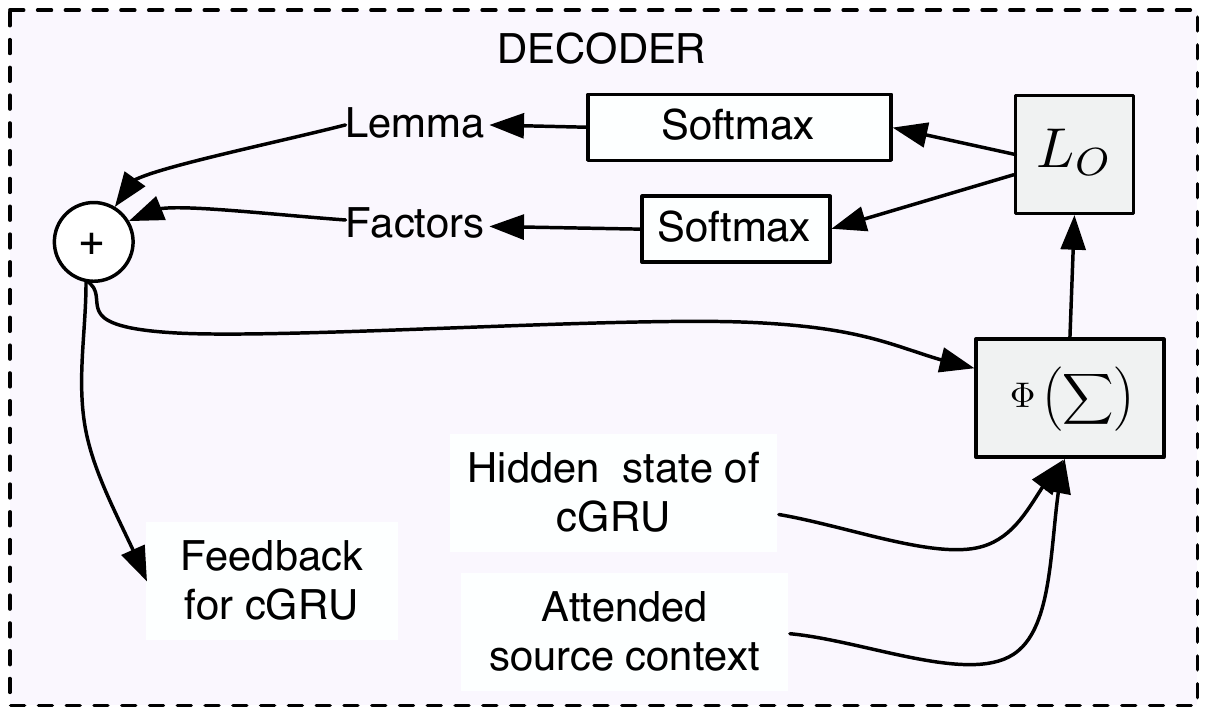}}
    \caption{\label{fi:factors_architecture} Detailed view of the decoder of the Factored NMT system}     
\end{figure}

For the sake of simplicity, only two symbols are generated: the lemma and the concatenation of the different factors (verb, tense, person, gender, number and case information).
The target words are then represented by a factored output: lemmas and factors.
Factors may help the translation process providing grammatical information to enrich the output.
The task of this work is English to French translation, English is a grammatically poor language and factors do not help for its translation, this has been tested in previous experiments. 
Therefore, we apply the factors only in the target side when translating to French which is a grammatically rich language.
In the example shown in Figure~\ref{fi:align_good}, from the verbal form in French \emph{devient}, we obtain the lemma \emph{devenir} and its factors \emph{VP3\#SL} (\textbf{V}erb, in \textbf{P}resent, \textbf{3}rd person, no gender (\textbf{\#}), \textbf{S}ingular and \textbf{L}owercased form).
Moreover, we can see the word \emph{inter\'essant} with the lemma \emph{inter\'essant} and factors \emph{Adj\#\#MSL} (\textbf{Adj}ective, no tense (\#) and no person (\#), \textbf{M}asculine gender, \textbf{S}ingular number and \textbf{L}owercased form).
The morphological analyser MACAON toolkit~\cite{macaon} is used to obtain the lemma and factors for each word taking into account its context with nearly 100\% accuracy.
The first entry is used in the few cases that MACAON proposes multiple words (e.g. same word written in two forms).

The FNMT architecture is presented in Figure~\ref{fi:factors_architecture}.
The encoder and attention mechanism of Figure~\ref{fi:nmt_archi} remain unchanged.
However, the decoder has been modified to get multiple outputs.
The hidden state of the conditional GRU (cGRU) is shared to produce simultaneously several outputs. 
The output from the layer $L_O$ has been diversified to two \emph{softmax} layers, one to generate the lemma and the other to generate the factors.
An additional design decision is related to the decoder feedback. 
Contrary to the word based model, where the feedback is naturally the previous word (see Figure~\ref{fi:nmt_archi}), we have multiple choices where multiple outputs are generated for each decoding time-step.
We have decided to use the concatenation of the embeddings of both generated symbols based on the work~\cite{Garcia16iwslt}.

The FNMT model may lead to sequences with a different length, since lemmas and factors are generated simultaneously but separately (each sequence ends after the generation of the end of sequence $<eos>$ token).
To avoid this, the sequences length is decided based on the lemmas stream length (i.e the length of the factors sequence is constrained to be equal to the length of the lemma sequence).
This is motivated by the fact that the lemmas contain most of the information of the final surface form (word). 

Once we obtain the factored outputs from the neural network, we need to combine them to obtain the surface form (word representation).
This operation is also performed with the MACAON tool, which, given a lemma and some factors, provides the word. 
Word forms given by MACAON toolkit have a 99\% success rate. 
In the cases (e.g. name entities) that the word corresponding to the lemma and factors is not found, the system outputs the lemma itself.

\section{Experiments}
\label{se:experiments}
 
We performed a set of experiments for Factored NMT (FNMT) and compared them with the word-based NMT and BPE-based NMT systems.

\subsection{Data processing and selection}

The systems are trained on the English to French (EN-FR) Spoken Language Translation task from IWSLT 2015 evaluation campaign\footnote{\small{https://sites.google.com/site/iwsltevaluation2015}}. 
We applied data selection using modified Moore-Lewis filtered by XenC~\cite{Rousseau} to obtain a sub part of the available parallel corpora (news-commentary, united-nations, europarl, wikipedia, and two crawled corpora).
The Technology Entertainment Design (TED)~\cite{cettoloEtAl:EAMT2012} corpus has been used as in-domain corpus. 

We preprocess the data to convert html entities and filter out the sentences with more than 50 words for both source and target languages.
Finally, we obtain a corpus of 2M sentences with 147k unique words for the English side and 266k unique words for the French side. French vocabulary is bigger than English 
since French is more highly inflected language. Table~\ref{tbl:data_stats} shows training, development and testing sets statistics.

\begin{table}[ht]
\centering
\caption{Datasets statistics}
\begin{tabular}{lccccc}
\toprule
\textbf{Data}  	& \textbf{Corpus name} 	& \textbf{Datasets} 	& \textbf{\# Sents}	& \textbf{\# Words EN-FR}   \\ \midrule
Training 		& train15 				& data selection 		& 2M              	& 147k-266k     \\
Development   	& dev15               	& dev10+test10+test13 	& 3.6k     			& 7.3k-8.9k		\\ 
Testing	 		& test15				& test11+test12 		& 1.9k 				& 4.5k-5.4k		\\ 
\bottomrule
\end{tabular}
\label{tbl:data_stats}
\end{table}

\subsection{Training}

Models are trained using NMTPY~\cite{nmtpy}, an NMT toolkit in Python based on Theano\footnote{ \small {\url{https://github.com/lium-lst/nmtpy}}}.
The following hyperparameters have been chosen to train the systems. 
The embedding and recurrent layers have the dimensions 620 and 1000, respectively.
The batch size is set to 80 sentences and the parameters are trained using the Adadelta~\cite{zeiler2012adadelta} optimizer. 
We clipped the norm of the gradient to be no more than 1~\cite{Pascanu} and initialize the weights using \emph{Xavier}~\cite{Glorot}.
The systems are validated on dev15 dataset using early stopping based on BLEU \cite{Papineni:2002:acl}. 
The vocabulary size of the source language is set to 30K. 
The output layer size of the baseline NMT system is set to 30K.  For the sake of comparability and consistency, 
the same value (30k) is used for the lemma output of the FNMT system. This 30K FNMT vocabulary includes 
17k lemmas obtained from the original NMT vocabulary (30k word level gives 17k lemmas when all the derived forms of the verbs, nouns, adjectives, etc are discarded) 
increased with additional new lemmas to fit the 30K desired value. The factors have 142 different units in their vocabulary.
When it comes to combining the lemmas and the factors vocabulary, the system is able to generate 172K different words, using the external linguistic resources, 
which is 6 times bigger than a standard word-based NMT vocabulary.\\

For BPE systems, bilingual vocabulary has been built using source and target language applying the joint vocabulary BPE approach.
In order to create comparable BPE systems, we set the number of merge operations for the BPE algorithm (the only hyperparameter of the method) as 30K minus the number of character according to the paper~\cite{Sennrich16BPE}. Then, we apply a total of 29388 merge operations to learn the BPE models on the training and validation sets.
During the decoding process, we use a beam size of 12 as used in~\cite{Bahdanau}.


\subsection{Quantitative results}
\label{sec:results}

The Factored NMT system aims at integrating linguistic knowledge into the decoder in order to overcome the restriction of having a large vocabulary at target side. 
We first compare our system with the standard word-level NMT system. 
For the sake of comparison with state of the art systems, we have built a subword system using the BPE method. 
Subwords were calculated at the input and the output side of the neural network as described in~\cite{Sennrich16BPE}.
The results are measured with two automatic metrics, the most common metric for machine translation BLEU and METEOR~\cite{Lavie:2007:acl}.
We evaluate on test15 dataset from the IWSLT 2015 campaign and results are presented in Table \ref{ta:factors_SOA_test_res}. 
 
\begin{table*}[!htbp]
\begin{center}
\caption{\label{ta:factors_SOA_test_res} Results on IWSLT test15. \%BLEU and \%METEOR performance of NMT and FNMT systems with and without UNK replacement (UR) are presented. 
For each system we provide the number of generated UNK tokens in the last column}
\begin{tabular}{ l|c|c|c|c|c } 
\toprule
     & \textbf {\%METEOR$\uparrow$} & \multicolumn{3}{|c|}{\textbf {\%BLEU$\uparrow$}}\\
\textbf{Model} & \textbf{word} & \textbf{word} & \textbf{lemma} &\textbf{factors} & \textbf{\#UNK} \\
   \hline
NMT / +UR	& 	62.21 / 63.38	& 41.80 / 42.74		& 45.10 	& 51.80 	& 1111\\ 
BPE					&	62.87	&	42.37			& 45.96		& 53.31		& 0	\\  
FNMT / +UR	&	\bf{64.10} / \bf{64.81}	& \bf 43.42 / 44.15		& \bf 47.18		& \bf54.24		& 604	\\ 
\bottomrule
 \end{tabular} 
\end{center}
\end{table*}

As we can see from the Table~\ref{ta:factors_SOA_test_res} results, the FNMT system obtains better \%BLEU and \%METEOR scores 
compared to the state of the art NMT and BPE systems. An improvement of about \textbf{1} \%BLEU point is achieved compared to the best baseline system (BPE). 
This improvement is even bigger (\textbf{1.4} \%BLEU point) when UNK replacement is applied to both systems.
In a quest to better understand the reasons of this improvement, we also computed the \%BLEU scores of each output level (lemmas and factors) for FNMT.
Theses scores are presented in Table~\ref{ta:factors_SOA_test_res}. 
The lemma and factors scores of NMT and BPE systems are obtained through a decomposing of their word level output into lemma and factors. 
We observe yet again that FNMT systems gives better score at both lemma and factors level. 
Replacement of unknown words has been performed using the alignments extracted from the attention mechanism. 
We have replaced the generated UNK tokens by translating its highest probability aligned source word.
We see an improvement of around 1 point \%BLEU score in both NMT and FNMT systems. \\

The last column of Table~\ref{ta:factors_SOA_test_res} shows, for each system, the the number of generated UNK tokens.
As shown in the table our FNMT system produces half of the UNK tokens compared to the word-based NMT system. 
This tends to prove that the Factored NMT system effectively succeed in modelling more words compared to the word 
based NMT system augmenting the generalization power of our model and preserving manageable output layer sizes.
Though we can see that BPE system does not produce UNK tokens, this is not reflected in the scores. 
Indeed, this can be due to the possibility of generation of incorrect words using BPE units in contrast to the FNMT system.



\subsection{Qualitative analysis}

\begin{figure}[!hbtp]
\centerline{\includegraphics[scale=0.5]{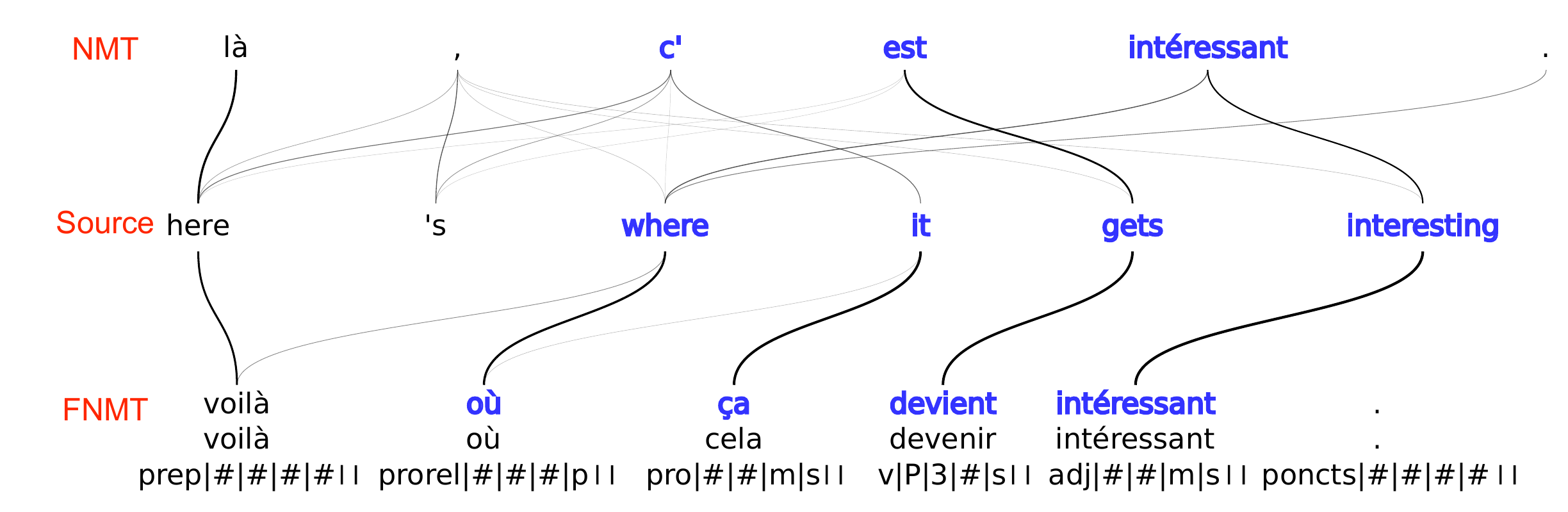}}
    \caption{\label{fi:align_good} Examples of NMT and FNMT outputs aligned against the source sentence }
\end{figure}

The strengths of FNMT are considered under this qualitative analysis.
We have studied and compared the translation outputs of NMT at word-level and BPE-level with the ones of FNMT systems. 
Two examples are presented in Figure~\ref{fi:align_good} and Table~\ref{ta:bpe_translation}.\\

The reference translation of the source sentence presented in Figure~\ref{fi:align_good} is \textit{``mais voilà où ça devient intéressant"}.
As we can see, contrary to the baseline NMT system, the FNMT system matches exactly the reference and thus produces the correct translation.
An additional interesting observation is that the alignment provided by the attention mechanism seems to be better defined and more helpful when using factors.
Also, one can notice the difference between the attention distributions made by the systems over the source sentence.
The NMT system first translated ``here'' into ``là'', added a coma, and then was in trouble for translating the rest of the sentence, which is reflected by the rather fuzzy attention weights. 
The FNMT system had better attention distribution over of the source sentence in this case.\\

\begin{table}[!hbtp]
\begin{center}
\setlength\tabcolsep{0.8pt}
\small
\caption{\label{ta:bpe_translation}Examples of translations with NMT, BPE and FNMT systems (without unknown words replacement)}
\begin{tabular}{ l|cccccccccccc } 
\toprule
\textbf{Src}	&  \multicolumn{10}{c}{we in medicine , I think , are \color{cyan}baffled}\\
\hline
\textbf{Ref} & \multicolumn{10}{c}{Je pense que en médecine nous sommes \color{cyan}dépassés}\\
\hline
\textbf{NMT} & Nous &, & en & médecine &, &je &pense &,& sont & \color{red}UNK\\
\hline
\textbf{BPE} &  nous &  , &  en &  médecine & , &  je &  pense &  , &  sommes &  \color{red}b@@ af@@ és\\
\hline
\textbf{FNMT} &  nous &  , &  en &  médecine &  , &  je &  pense &  , &  sont &  \color{blue}déconcertés\\
Lemmas &  lui &  , &  en &  médecine &  , &  je &  penser &  , &  être &  \color{blue}déconcerter\\
Factors &  pro-1-p-l &  pct-l &  prep-l &  nc-f-s-l &  pct-l &  cln-1-s-l &  v-P-1-s-l &  pct-l &  \color{red}{v-P-3-p-l} &  \color{blue}vppart-K-m-p-l  \\
\bottomrule
 \end{tabular} 
\end{center}
\vspace{-0.3cm}
\end{table}

Table~\ref{ta:bpe_translation} shows another example comparing NMT, BPE and FNMT systems.
The NMT system generated an unknown token (UNK) when translating the English word \textit{``baffled"}.
We observe that BPE translates \textit{``baffled"} to \textit{``bafés"} which does not exist in French.
This error probably comes from the shared vocabulary between the source and target languages creating an incorrect word very similar to its aligned source tokens.
FNMT translates it to \textit{``déconcertés"} which is a better translation than in the reference. One should note that it is not generated by the unknown word replacement method. 
However, for this particular example, an error on the factors leads to the word \textit{``sont"} instead of \textit{``sommes"}, resulting in lower automatic scores for FNMT output.

\section{Conclusion}
\label{se:conclusion}

In this paper, the Factored NMT approach has been further explored.
Factors based on linguistic \textit{a priori} knowledge have been used to decompose the target words.
This approach outperforms a strong baseline system using subword units computed with byte pair encoding.
Our FNMT system is able to model an almost 6 times bigger word vocabulary with only a slight increase of the computational cost.
By these means, the FNMT system is able to halve the generation of unknown tokens compared to word-level NMT. 
Using a simple unknown word replacement procedure involving a bilingual dictionary, we are able to obtain even better results (+0.8 \%BLEU compared to previous best system).\\

Also, the use of external linguistic resources allows us to generate new word forms that would not be included in the standard NMT system \emph{shortlist}.
The advantage of this approach is that the new generated words are controlled by the linguistic knowledge, that avoid producing incorrect words, as opposed to actual systems using BPE.
We demonstrated the performance of such a system on an inflected language (French). 
The results are very promising for use with highly inflected languages like Arabic or Czech.\\

\section*{Acknowledgments}
This work was partially funded by the French National Research Agency (ANR) through the CHIST-ERA M2CR project, under the contract number ANR-15-CHR2-0006-01.

\bibliography{fnmt}

\begin{thebibliography}{10}
\providecommand{\url}[1]{\texttt{#1}}
\providecommand{\urlprefix}{URL }

\bibitem{Alexandrescu}
Alexandrescu, A.: Factored neural language models. In: In HLT-NAACL (2006)

\bibitem{Bahdanau}
Bahdanau, D., Cho, K., Bengio, Y.: Neural machine translation by jointly
  learning to align and translate. CoRR  abs/1409.0473 (2014)

\bibitem{nmtpy}
Caglayan, O., Garc\'{i}a-Mart\'{i}nez, M., Bardet, A., Aransa, W., Bougares,
  F., Barrault, L.: Nmtpy: A flexible toolkit for advanced neural machine
  translation systems. arXiv preprint arXiv:1706.00457  (2017),
  \url{http://arxiv.org/abs/1706.00457}

\bibitem{cettoloEtAl:EAMT2012}
Cettolo, M., Girardi, C., Federico, M.: Wit$^3$: Web inventory of transcribed
  and translated talks. In: Proceedings of the 16$^{th}$ Conference of the
  European Association for Machine Translation (EAMT). pp. 261--268. Trento,
  Italy (May 2012)

\bibitem{Cho}
Cho, K., van Merrienboer, B., G{\"{u}}l{\c{c}}ehre, {\c{C}}., Bougares, F.,
  Schwenk, H., Bengio, Y.: Learning phrase representations using {RNN}
  encoder-decoder for statistical machine translation. CoRR  abs/1406.1078
  (2014)

\bibitem{Chung}
Chung, J., Cho, K., Bengio, Y.: A character-level decoder without explicit
  segmentation for neural machine translation. CoRR  abs/1603.06147 (2016)

\bibitem{Costa-Jussa}
Costa{-}Juss{\`{a}}, M.R., Fonollosa, J.A.R.: Character-based neural machine
  translation. CoRR  abs/1603.00810 (2016)

\bibitem{cgru}
Firat, O., Cho, K.: Conditional gated recurrent unit with attention mechanism.
  \url{github.com/nyu-dl/dl4mt-tutorial/blob/master/docs/cgru.pdf} (2016)

\bibitem{Garcia16iwslt}
Garc\'ia-Mart\'inez, M., Barrault, L., Bougares, F.: Factored neural machine
  translation architectures. In: Proceedings of the International Workshop on
  Spoken Language Translation. IWSLT'16, Seattle, USA (2016),
  \url{http://workshop2016.iwslt.org/downloads/IWSLT\_2016\_paper\_2.pdf}

\bibitem{Glorot}
Glorot, X., Bengio, Y.: Understanding the difficulty of training deep
  feedforward neural networks. In: In Proceedings of the International
  Conference on Artificial Intelligence and Statistics (AISTATS’10). Society
  for Artificial Intelligence and Statistics (2010)

\bibitem{Jean}
Jean, S., Cho, K., Memisevic, R., Bengio, Y.: On using very large target
  vocabulary for neural machine translation. CoRR  abs/1412.2007 (2014)

\bibitem{Koehn}
Koehn, P., Hoang, H., Birch, A., Callison-Burch, C., Federico, M., Bertoldi,
  N., Cowan, B., Shen, W., Moran, C., Zens, R., Dyer, C., Bojar, O.,
  Constantin, A., Herbst, E.: Moses: Open source toolkit for statistical
  machine translation. In: Proceedings of the 45th Annual Meeting of the ACL on
  Interactive Poster and Demonstration Sessions. pp. 177--180. ACL '07,
  Association for Computational Linguistics, Stroudsburg, PA, USA (2007)

\bibitem{Lavie:2007:acl}
Lavie, A., Agarwal, A.: Meteor: an automatic metric for mt evaluation with high
  levels of correlation with human judgments. In: Proceedings of the Second
  Workshop on Statistical Machine Translation. pp. 228--231. StatMT '07,
  Association for Computational Linguistics, Stroudsburg, PA, USA (2007)

\bibitem{Le}
Le, H.S., Oparin, I., Messaoudi, A., Allauzen, A., Gauvain, J.L., Yvon, F.:
  Large vocabulary {SOUL} neural network language models. In: INTERSPEECH
  (2011), \url{sources/Le11large.pdf}

\bibitem{Ling15}
Ling, W., Trancoso, I., Dyer, C., Black, A.W.: Character-based neural machine
  translation. CoRR  abs/1511.04586 (2015)

\bibitem{LuongSLVZ14}
Luong, T., Sutskever, I., Le, Q.V., Vinyals, O., Zaremba, W.: Addressing the
  rare word problem in neural machine translation. CoRR  abs/1410.8206 (2014),
  \url{http://arxiv.org/abs/1410.8206}

\bibitem{macaon}
Nasr, A., Béchet, F., Rey, J.F., Favre, B., Roux, J.L.: Macaon, an nlp tool
  suite for processing word lattices. In: Proceedings of the ACL-HLT 2011
  System Demonstrations. pp. 86--91 (2011)

\bibitem{niehues}
Niehues, J., Ha, T.L., Cho, E., Waibel, A.: Using factored word representation
  in neural network language models. In: Proceedings of the First Conference on
  Machine Translation. pp. 74--82. Association for Computational Linguistics,
  Berlin, Germany (August 2016)

\bibitem{Papineni:2002:acl}
Papineni, K., Roukos, S., Ward, T., Zhu, W.J.: Bleu: a method for automatic
  evaluation of machine translation. In: Proceedings of the 40th Annual Meeting
  on Association for Computational Linguistics. pp. 311--318. ACL '02,
  Stroudsburg, PA, USA (2002)

\bibitem{Pascanu}
Pascanu, R., Mikolov, T., Bengio, Y.: Understanding the exploding gradient
  problem. CoRR  abs/1211.5063 (2012)

\bibitem{Rousseau}
Rousseau, A.: {XenC}: An open-source tool for data selection in natural
  language processing. The Prague Bulletin of Mathematical Linguistics  100,
  73--82 (2013)

\bibitem{Sennrich2016HowGI}
Sennrich, R.: How grammatical is character-level neural machine translation?
  assessing mt quality with contrastive translation pairs. CoRR  abs/1612.04629
  (2016)

\bibitem{SennrichH16}
Sennrich, R., Haddow, B.: Linguistic input features improve neural machine
  translation. CoRR  abs/1606.02892 (2016)

\bibitem{Sennrich16BPE}
Sennrich, R., Haddow, B., Birch, A.: Neural machine translation of rare words
  with subword units. In: Proceedings of the 54th Annual Meeting of the
  Association for Computational Linguistics (Volume 1: Long Papers). pp.
  1715--1725. Association for Computational Linguistics (2016)

\bibitem{SutskeverVL14}
Sutskever, I., Vinyals, O., Le, Q.V.: Sequence to sequence learning with neural
  networks. CoRR  abs/1409.3215 (2014)

\bibitem{Wu2012FactoredRN}
Wu, Y., Yamamoto, H., Lu, X., Matsuda, S., Hori, C., Kashioka, H.: Factored
  recurrent neural network language model in ted lecture transcription. In:
  IWSLT (2012)

\bibitem{zeiler2012adadelta}
Zeiler, M.D.: Adadelta: an adaptive learning rate method. arXiv preprint
  arXiv:1212.5701  (2012)

\end{thebibliography}
\bibliographystyle{splncs03}
\appendix

\end{document}